\documentclass{article}

\PassOptionsToPackage{numbers, sort&compress}{natbib}
\usepackage[preprint]{neurips_2023}

\usepackage[utf8]{inputenc} % allow utf-8 input
\usepackage[T1]{fontenc}    % use 8-bit T1 fonts
\usepackage{url}            % simple URL typesetting
\usepackage{booktabs}       % professional-quality tables
\usepackage{amsfonts}       % blackboard math symbols
\usepackage{nicefrac}       % compact symbols for 1/2, etc.
\usepackage{microtype}      % microtypography
\usepackage{enumitem}
\usepackage{wrapfig} 
\usepackage{graphicx}
\usepackage{subfigure}
\usepackage{amsmath}
\usepackage{amssymb}
\usepackage{booktabs}
\usepackage{bbm}
\usepackage[dvipsnames,svgnames,x11names,table]{xcolor} % colors
\usepackage{pifont}
\usepackage{bbding}

\usepackage{wrapfig}
\usepackage{multirow}
\usepackage{transparent}
\usepackage{tikz}

\definecolor{pretty-blue}{RGB}{0, 113, 188}
\definecolor{brown}{RGB}{201, 104, 71}

\usepackage[colorlinks=True,
            linkcolor=blue,
            anchorcolor=blue,  
            pagebackref,
            urlcolor=blue,
            citecolor=teal,
            ]{hyperref}

\title{\ \ \ \ \ \ Vary: Scaling up the Vision Vocabulary for \\ \ \ \ \ Large Vision-Language Models}

\author{
Haoran Wei$^{1}$\thanks{Equal contribution}~,
Lingyu Kong$^{2*}$, Jinyue Chen$^{2}$, Liang Zhao$^{1}$,  \bf{Zheng Ge}$^{1}$\thanks{Project leader}~, \\ \bf{Jinrong Yang}$^{3}$, \bf{Jianjian Sun}$^{1}$, \bf{Chunrui Han}$^{1}$, Xiangyu Zhang$^{1}$\\
$^{1}$MEGVII Technology \ \  $^{2}$University of Chinese Academy of Sciences  \\ $^{3}$Huazhong University of Science and Technology\\
{\url{https://varybase.github.io/}}
}

\begin{document}

\maketitle

\begin{tikzpicture}[remember picture,overlay,shift={(current page.north west)}]
\node[anchor=north west, xshift=3.7cm, yshift=-3.2cm]{\scalebox{-1}[1]{\includegraphics[width=1.50cm]{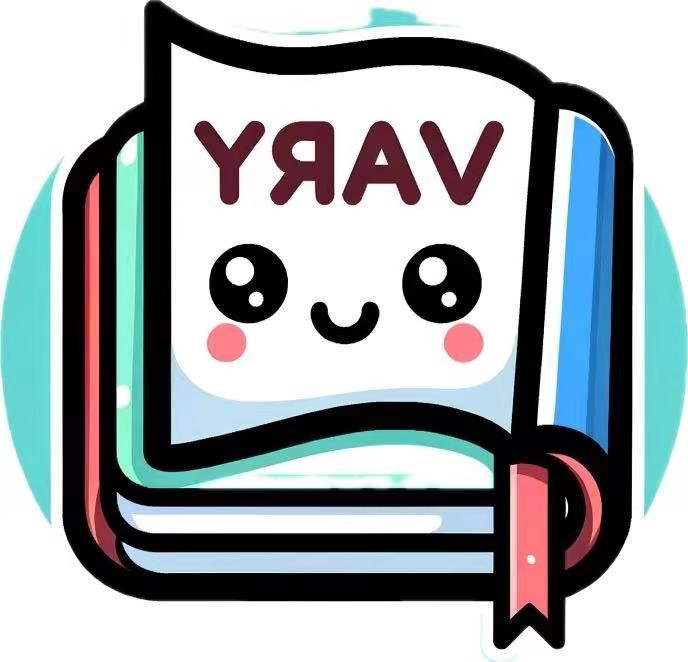}}};
\end{tikzpicture}

% \begin{tikzpicture}[remember picture,overlay,shift={(current page.north west)}]
% \node[anchor=north west, xshift=3.0cm, yshift=-3.0cm]{\scalebox{-1}[1]{\includegraphics[width=1.5cm]{Figs/logo.png}}};
% \end{tikzpicture}

% \emph{“Beauty and brains, pleasure and usability – they should go hand in hand.”}

% \rightline{---Donald A. Norman}

\begin{abstract}
Modern Large Vision-Language Models (LVLMs) enjoy the same vision vocabulary -- CLIP, which can cover most common vision tasks. However, for some special vision task that needs dense and fine-grained vision perception, \textit{e.g.}, document-level OCR or chart understanding, especially in non-English scenarios, the CLIP-style vocabulary may encounter low efficiency in tokenizing the vision knowledge and even suffer out-of-vocabulary problem. Accordingly, we propose \textbf{Vary}, an efficient and effective method to scale up the  \textbf{V}ision vocabul\textbf{ary} of LVLMs. The procedures of Vary are naturally divided into two folds: the generation and integration of a new vision vocabulary. In the first phase, we devise a vocabulary network along with a tiny decoder-only transformer to produce the desired vocabulary via autoregression. In the next, we scale up the vanilla vision vocabulary by merging the new one with the original one (CLIP), enabling the LVLMs can quickly garner new features. Compared to the popular BLIP-2, MiniGPT4, and LLaVA, Vary can maintain its vanilla capabilities while enjoying more excellent fine-grained perception and understanding ability. Specifically, Vary is competent in new document parsing features (OCR or markdown conversion) while achieving 78.2\% ANLS in DocVQA and 36.2\% in MMVet. Our code will be publicly available on the homepage.

\end{abstract}

\section{Introduction}
\label{intro}

Recently, research into vision dialogue robots~\cite{BLIP2,Flamingo,llava,minigpt4,InstructGPT} has been gaining significant traction. These human-like models, mainly relying on two components (large language models (LLMs)~\cite{GPT-2,GPT3,OPT,llama,GPT4} and vision vocabulary networks), can not only converse based on user's input image but also perform well on simple downstream tasks, such as VQA~\cite{COCO,TextVQA}, Image caption~\cite{coco_text}, OCR~\cite{OCRVQA}, and so on. Hence, it is undeniable that large vision-language models (LVLMs) are driving the AI community towards the direction of artificial general intelligence (AGI).

Popular GPT-4~\cite{GPT4}-like LVLMs, \textit{e.g.}, BLIP2~\cite{BLIP2}, MiniGPT4~\cite{minigpt4},LLaVA~\cite{llava}, Qwen-VL~\cite{Qwen-VL}, and \textit{etc}.~\cite{dong2023dreamllm,zhao2023chatspot,yu2023merlin} enjoy a stunning performance in multiple aspects with their own programming paradigm: Based on an LLM~\cite{OPT,T5}, BLIP-2 proposes the Q-former, a BERT~\cite{Bert} like network as a vision input embedding layer, aiming to align the image tokens to a text special. Inherited the structure of BLIP-2, MiniGPT-4 introduces 3500 high-quality image-text pairs as self-supervised fine-tuning (SFT) data, allowing it can ``talk'' like GPT-4. Unlike BLIP-2, LLaVA utilizes a linear layer as the vision embedding layer, which is similar with the text input embedding layer in the text tokenizer, ensuring the consistency in the structure of image and text branches. For Qwen-VL, it utilizes a cross-attention layer to sample and align the image tokens, making the model can accept larger input resolution.  Although the above LVLMs' vision input embedding networks are variable (\textit{e.g.}, MLP, Qformer, Perceiver~\cite{Flamingo}), their vision vocabulary is almost identical (a CLIP-based~\cite{radford2021learning} VIT) which we argue maybe a bottle-neck.

\begin{figure}[t]
\centering
\includegraphics[width=1.0\textwidth]{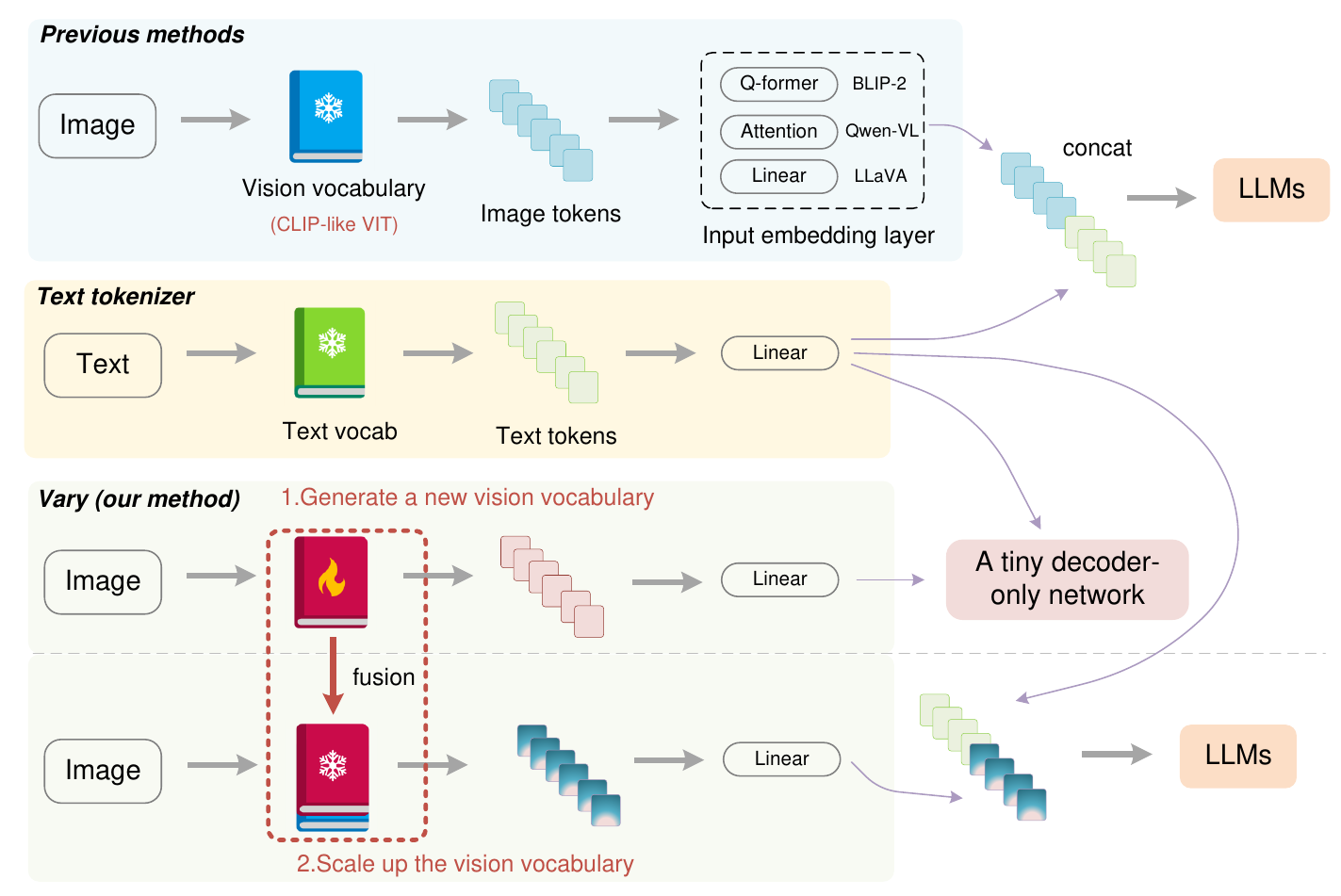}
\caption{Previous method \textit{vs.} Vary: Unlike other models that use a ready-made vision vocabulary, the processes of Vary can be divided into two stages: the generation and fusion of vision vocabulary. In the first stage, we use a ``vocabulary network'' along with a tiny decoder-only network to produce a powerful new vision vocabulary via auto-regression. In the second stage, we fuse the vision vocabulary with the original one to provide new features for the LVLMs efficiently.}
\label{fig1}
\end{figure}

It is recognized that CLIP-VIT is a tremendous general vision vocabulary, which is trained via contrastive learning upon more than 400M~\cite{schuhmann2021laion} image-text pairs, covering most natural images and vision tasks. However, for some special scenarios, \textit{e.g.}, high-resolution perception, Non-English OCR, Document/Chart understanding, and so on, the CLIP-VIT may regard them as a ``foreign language'', leading to inefficient tokenizing, \textit{i.e.}, difficulty in encoding all vision information into a fixed number (usually 256) of tokens. Although mPlug-Owl~\cite{ye2023mplug} and Qwen-VL alleviate the above issues by unfreeze its vision vocabulary network (a CLIP-L or CLIP-G), we argue that such manner may not be reasonable due to three aspects: 1) it may overwrite the knowledge of the original vocabulary; 2) the training efficiency of updating a vision vocabulary upon a relative large LLM (7B) is low; 3) it can not allow the vision vocabulary network to ``see'' an image multiple times (train a dataset with multiple epochs) due to the strong memory ability of LLMs. Therefore, a natural question is: \textit{Is there a strategy that can simplify and effectively intensify the visual vocabulary?}

In this paper, we propose Vary, an efficient and user-friendly approach, to answer the above question. Vary is inspired by the text vocabulary expansion manner in vanilla LLMs~\cite{vicuna}, \textit{i.e.}, when transferring an English LLM to another foreign language, such as Chinese, it's necessary to expand the text vocabulary to lift the encoding efficiency and model performance under the new language. Intuitively, for the vision branch, if we feed the ``foreign language'' image to the model, we also need to scale up the vision vocabulary. In Vary, the process of vocabulary scaling up can be divided into two steps: 1) generate a new vision vocabulary that can make up the old one (CLIP); 2) integrate the new and old vocabularies. As shown in Figure~\ref{fig1}, we build a small-size pipeline which is consisting of a vocabulary network and a tiny decoder-only transformer in the first step to train the vocabulary model via predicting the next token. It is worth noting that the autoregressive-based process of generating a vocabulary is perhaps more suitable for dense perception tasks than that based on contrastive learning like CLIP. On the one hand, the next-token way can allow the vision vocabulary to compress longer texts. On the other hand, the data formats that can be used in this manner are more diverse, such as VQA~\cite{STVQA,DocVQA} data with prompt. After preparing the new vision vocabulary, we add it to the vanilla LVLMs to introduce new features. In this process, we freeze both the new and old vocabularies networks to avoid the visual knowledge being overwritten. 

Afterward scaling up the vision vocabulary, our LVLM can achieve more fine-grained vision perception, such as document-level Chinese/English OCR, book image to markdown or \textit{\LaTeX{}}, Chinese/English chart understanding, and so on, while ensuring its original capabilities (conversation, VQA, caption, \textit{etc}.). Besides, we provide methods for producing synthetic data and validate its importance in document/chart understanding. More importantly, Vary is a useful strategy to strengthen the visual vocabulary of LVLMs, which can be utilized at arbitrary downstream visual tasks that CLIP is not good at. In addition to the document and chart parsing mentioned in this paper, we believe that Vary still enjoys more fine-grained tasks and we appeal to researchers to rethink the design ideas of LVLMs from the perspective of visual vocabulary construction.

\section{Related Works}

\subsection{Large Language Models}
Over the past year, significant attention has been drawn to large language models (LLMs) in the fields of both natural language processing (NLP) and computer vision (CV). This heightened attention stems from LLMs' outstanding performance in diverse aspects, especially the powerful world knowledge base and universal capabilities. Current LLMs enjoy a unified transformer architecture which is exemplified by BERT~\cite{Bert}, GPT-2~\cite{GPT-2}, T5~\cite{T5}, \textit{etc}. Subsequently, researchers have uncovered the concept of an "emergent ability"~\cite{wei2022emergent} in LLMs. This implies that as language model sizes reach a certain threshold, there may be a qualitative leap in their capabilities. Furthermore, InstructGPT~\cite{InstructGPT} and ChatGPT~\cite{ChatGPT} find that Reinforcement Learning with Human Feedback (RLHF)~\cite{christiano2017deep} can further lift the performance of the "talk robot''. Motivated by the tremendous success of the GPT series, a multitude of other open-source LLMs have emerged, including OPT~\cite{OPT}, LLaMA~\cite{llama}, GLM~\cite{GLM}, and so on. Building upon these openly available LLMs, numerous tailored fine-tuned models have been introduced to develop LLMs for diverse applications, especially LLaMA-driven models,\textit{e.g.}, Alphaca~\cite{alpaca}, Vicuna~\cite{vicuna}, which have become the de-facto component for a Large Vision-Language Model (LVLM).

\subsection{LLM-based Large Vision-Language Models}
LLM's robust zero-shot capabilities and logical reasoning make it play the central controller role within an LVLM. There are two primary pipeline styles: plugin-based and end-to-end model. Plugin-based methods~\cite{VisualChatGPT, MMREACT, Hugginggpt, taskmatrix, yang2023gpt4tools} typically regard LLMs as an agent to invoke various plugins from other foundational or expert models, executing specific functions in response to human instructions. While such methods offer versatility, they have limitations in terms of plugin invocation efficiency and performance. Conversely, end-to-end LVLMs usually rely on a single large multimodal model to facilitate interactions. Following this approach, Flamingo~\cite{Flamingo} introduces a gated cross-attention mechanism trained on billions of image-text pairs to align vision and language modalities, demonstrating strong performance in few-shot learning. BLIP-2~\cite{BLIP2} introduces Q-Former to enhance the alignment of visual features with the language space. More recently, LLaVA~\cite{llava} proposes using a simple linear layer to replace Q-Former and designed a two-stage instruction-tuning procedure. 

Despite the remarkable performance of existing methods, they are confined to the same and limited vision vocabulary -- CLIP-VIT~\cite{radford2021learning}. For an LVLM, CLIP-VIT is a tremendous general vision vocabulary that is trained via contrastive learning upon million-level image-texts pairs, which can cover most nature images and vision tasks, \textit{e.g.}, VQA, Caption, Easy English OCR. However, some images under special scenarios, \textit{e.g.}, high-resolution image, Non-English OCR, Document/Chart understanding, and so on, will still be regarded as a ``foreign language'' by CLIP-VIT, leading to vision out-of-vocabulary problem, which will in turn become a bottleneck for LVLMs.

\section{Method}
\label{methods}
% In this section, 
\subsection{Architecture}

\begin{figure}[t]
\centering
\includegraphics[width=1.0\textwidth]{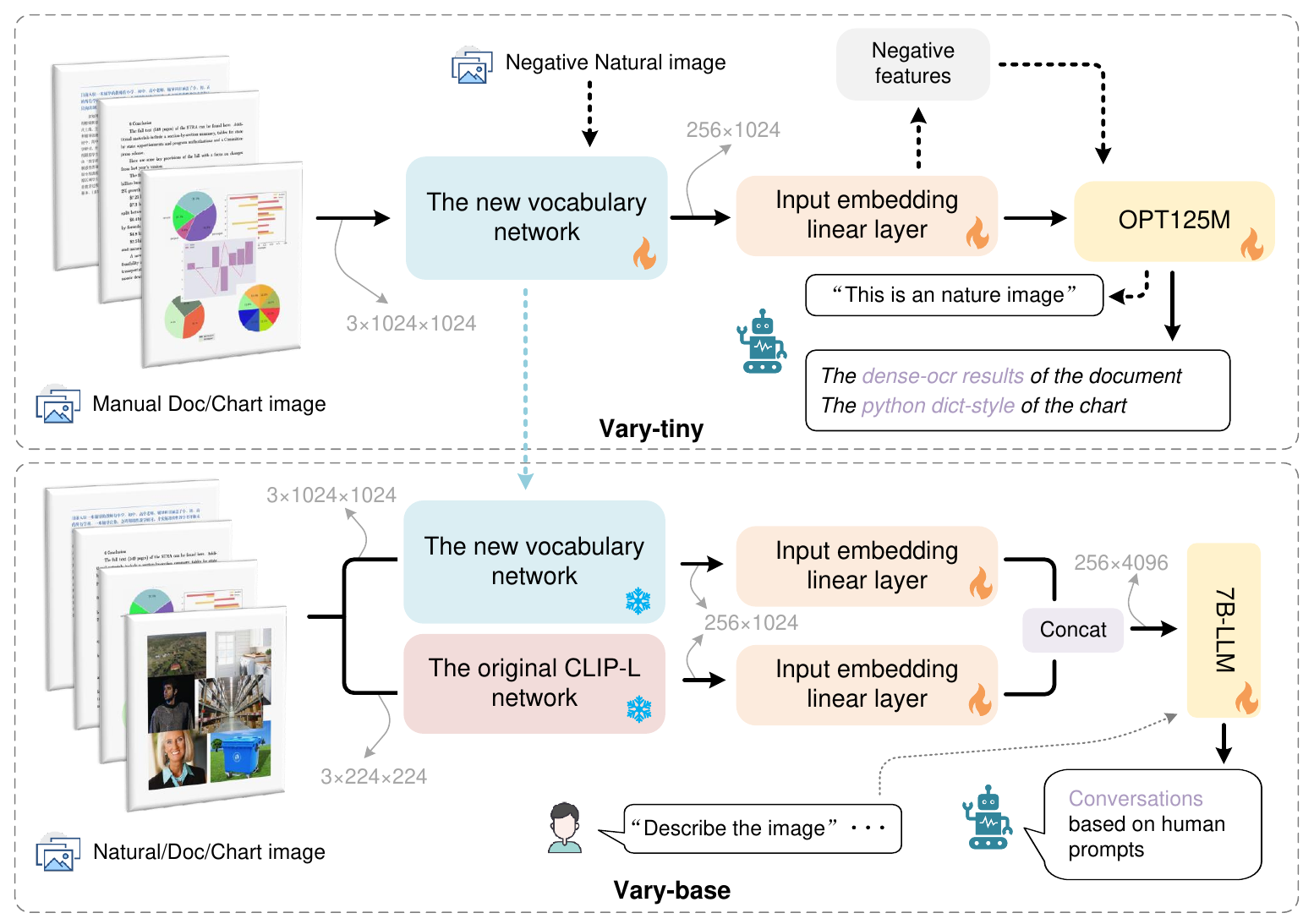}
\caption{Overview of the Vary. There are two types of Vary form: Vary-tiny and Vary-base. Vary-tiny is mainly focused on generating a new vision vocabulary while Vary-base is our new LVLM aiming to handle various visual tasks based on the new vision vocabulary. }
\label{fig2}
\end{figure}

Vary enjoys two conformations: Vary-tiny and Vary-base, as shown in Figure~\ref{fig2}. We devise the Vary-tiny to ``write'' a new vision vocabulary and the Vary-base to make use of the new vocabulary. Specifically, Vary-tiny is mainly composed of a vocabulary network and a tiny OPT-125M~\cite{OPT}. Between the two modules, we add a linear layer to align the channel dimensions. There is no text input branch in Vary-tiny due to it is a primary focus on fine-grained perception. We hope the new vision vocabulary network can excel in processing artificial images, \textit{i.e.}, documents, and charts, to compensate for CLIP's shortcomings. At the same time, we also expect that it will not be a noise for CLIP when tokenizing natural images. Accordingly, during generating, we feed the manual document and chart data as positive samples and natural images as negatives to train Vary-tiny. 
After completing the above process, we extract the vocabulary network and add it to a large model to build the Vary-base. As shown in the lower half of Figure~\ref{fig2}, the new and old vocabulary networks enjoy independent input embedding layers and are integrated before the LLM. In such a stage, we freeze both weights of new and old vision vocabulary networks and unfreeze the weights of other modules.

\subsection{Towards Generating a New Vision Vocabulary}
\subsubsection{The new vocabulary network}

\begin{wrapfigure}{r}{0.5\textwidth}
\vspace{-5ex}
	\centering
	\includegraphics[width=0.5\textwidth]{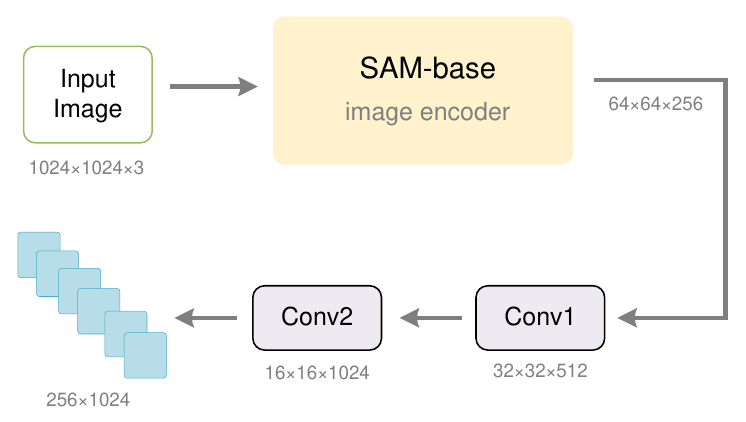}
	\caption{The structure of new vision vocabulary network. We add two convolution layers to convert the output to be similar with CLIP. }
	\label{fig3}
\vspace{-3ex}
\end{wrapfigure}

We use the SAM~\cite{kirillov2023segment} pretrained ViTDet~\cite{li2022exploring} image encoder (base scale) as the main part of the new vocabulary network of Vary. Due to the input resolution of the SAM-base is (1024$\times$1024) while the output stride is 16, the feature shape of the last layer is (64$\times$64$\times$256 for H$\times$W$\times$C) that can not be aligned to the output of CLIP-L (256$\times$1024 for N$\times$C). Hence, we add two convolution layers, which we found is a good token merging unit, behind the last layer of the SAM initialized network, as shown in Figure~\ref{fig3}. The first convolution layer possesses a kernel size of 3, aiming to transfer the feature shape to 32$\times$32$\times$512. The setting of the second conv layer is the same as the first one, which can further convert the output shape to 16$\times$16$\times$1024. After that, we flattened the output feature to 256$\times$1024 to align the image token shape of CLIP-VIT. 

\subsubsection{Data engine in the generating phrase} \label{data1}
\textbf{Documnet data.} We select the high-resolution document image-text pairs as the main positive dataset used for the new vision vocabulary pretrain due to the dense OCR can effectively validate the fine-grained image perception ability of the model. To our knowledge, there is no publicly available dataset of English and Chinese documents, so we create our own.
We first collect pdf-style documents from open-access articles on arXiv and CC-MAIN-2021-31-PDF-UNTRUNCATED for the English part and collect from e-books on the Internet for the Chinese part. Then we use \textit{fitz} of PyMuPDF to extract the text information in each pdf page and convert each page into a PNG image via \textit{pdf2image} at the same time. During this process, we construct 1M Chinese and  1M English document image-text pairs for training.  

% The text extracted using the underlying library has no formatting (markdown or \textit{\LaTeX{}}). 

% \item \textit{\LaTeX{}} rendering. The We collect English and Chinese text corpus online and created the document image-text pairs through \textit{\LaTeX{}} rendering. We select 10 templates and lots of mathematical formulas to render images in a text and formula interleaved form. Besides, we transfer the text ground-truth to a markdown format. By this construction process, there were 0.9 million English pages and 0.4 million Chinese pages generated.
% \end{itemize}

\textbf{Chart data.} We find current LVLMs are not good at chart understanding, especially Chinese charts, so we choose it as another main knowledge that needs to be ``written'' into the new vocabulary.  For chart image-text pair, we all follow the rendering way. We select both the \textit{matplotlib} and \textit{pyecharts} as the rendering tools.  For matplotlib-style chart, we built 250k in both Chinese and English. While for pyecharts, we build 500k for both Chinese and English. Besides, we convert the text ground truth of each chart to a python-dict form. The texts used in the chart, \textit{e.g.}, title, x-axis, and y-axis, are randomly selected from the Natural Language Processing (NLP) corpus downloaded from the Internet.

\textbf{Negative natural image.} For natural image data that CLIP-VIT is good at, we need to ensure that the newly introduced vocabulary does not cause noise. Consequently, we construct negative natural image-text pairs to enable the new vocabulary network to encode correctly when seeing natural images. We extract 120k images in the COCO~\cite{COCO} dataset with each image corresponding to a text. The text part is randomly selected from follows sentences: "It's an image of nature"; "Here's a nature picture";  "It's a nature photo"; "This is a natural image"; "That's a shot from nature".

% \begin{figure}[t]
% \centering
% \includegraphics[width=1.0\textwidth]{Figs/docdata.png}
% \caption{Pipeline of generating doc-text pairs. (a) Sample corpus from the Internet. (b) Random sample some formulas from the arXiv tex and mix with corpus. (c) Parser mixed corpus into latex format. (d) Randomly select a \LaTeX template to compile pdf pages. (e) Parser mixed corpus into markdown format. (f) Save markdown content into JSON format.}
% \label{fig-doc}
% \end{figure}

\subsubsection{Input format} 

We train all parameters of the Vary-tiny with image-text pairs by autoregression. The input format follows popular LVLMs~\cite{KOSMOS}, \textit{i.e}, the image tokens are packed with text tokens in the form of a prefix. Specifically, we use two special tokens "<img>" and "</img>" to indicate the position of the image tokens as the input of an interpolated OPT-125M (4096 tokens). During training, the output of Vary-tiny is only text, and "</s>" is regarded as the \textit{eos} token.

\begin{figure}[t]
\centering
\includegraphics[width=1.0\textwidth]{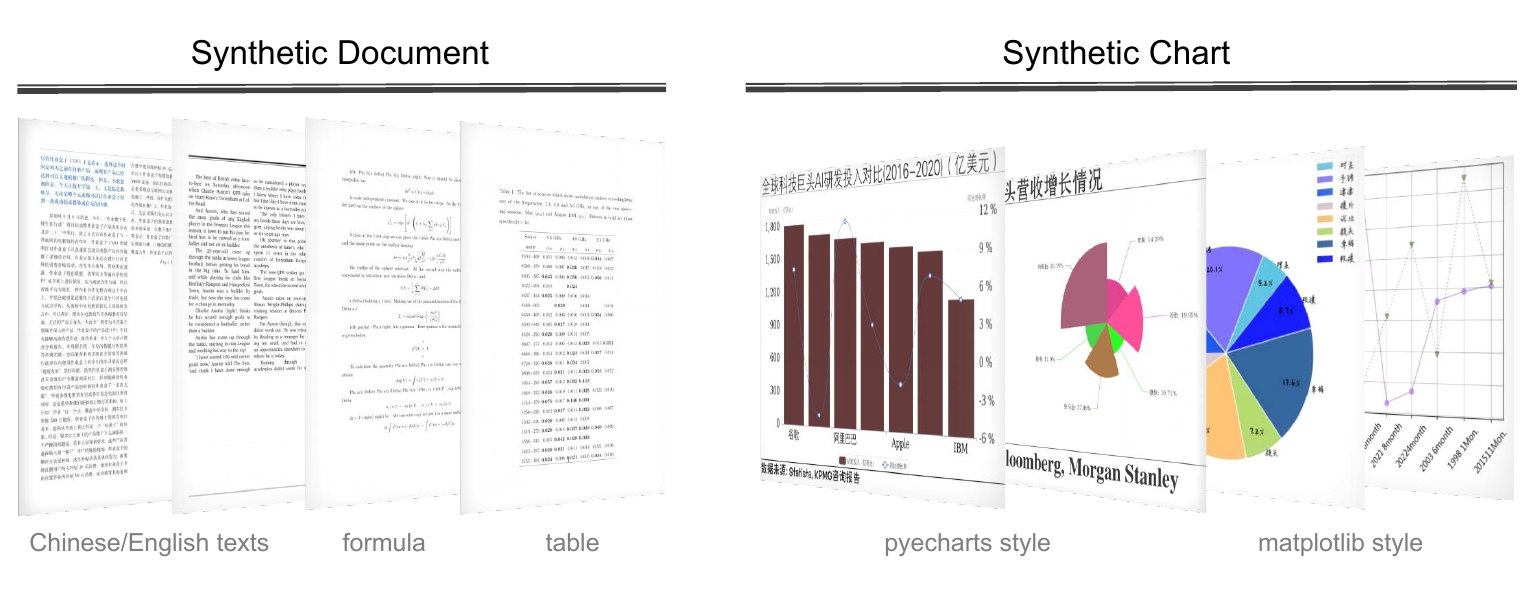}
\caption{Visualization of synthetic data. We use \textit{pdflatex} to render documents and utilize \textit{pyecharts/matplotlib} to render charts. Document data obtains Chinese/English texts, formulas, and tables. Chart data includes Chinese/English bar, line, pie, and composite styles.}
\label{fig4}
\end{figure}

\subsection{Towards Scaling Up the Vision Vocabulary}

\subsubsection{The structure of Vary-base} 
After completing the training of the vocabulary network, we introduce it to our LVLM -- Vary-base. Specifically, we parallelize the new vision vocabulary with the original CLIP-VIT. Both two vision vocabularies enjoy an individual input embedding layer, \textit{i.e.}, a simple linear. As shown in Figure~\ref{fig2}, the input channel of the linear is 1024 and the output is 2048, ensuring the channel of image tokens after concatenating is 4096, which exactly aligns the input of LLM (Qwen-7B~\cite{qwen} or Vicuna-7B~\cite{vicuna}).

\subsubsection{Data engine in the scaling up phrase} 

\textbf{\textit{\LaTeX{}} rendering document}. Except for the collecting document data in Section~\ref{data1}, we also need data to enjoy some format, \textit{e.g.}, supporting formula, and table. To this end, we create document data through \textit{\LaTeX{}} rendering.  Firstly, we collected some \textit{.tex} source files on arxiv, and then extracted tables, mathematical formulas, and plain texts using regular expressions. Finally, we re-render these contents with the new template we prepared by \textit{pdflatex}. We collect 10+ templates to perform batch rendering. Besides, we transfer the text ground truth of each document page to a \textit{mathpix} markdown style to unify the format. By this construction process, we acquired 0.5 million English pages and 0.4 million Chinese pages. Some samples are shown in Figure~\ref{fig4}.

\textbf{Semantic association chart rendering}.  In Section~\ref{data1}, we batch render chart data to train the new vocabulary network. However, the texts (title, x-axis values, and y-axis values) in those rendered charts suffer low correlation because they are randomly generated. This issue is not a problem in the vocabulary-generating process as we only hope that the new vocabulary can efficiently compress visual information. However, in the training stage of the Vary-base, due to unfreezing the LLM, we hope to use higher quality (strongly correlated content) data for training. Therefore, we use GPT-4~\cite{GPT4} to generate some charts using relevant corpus and then we utilize the high-quality corpus to addition render 200k chart data for the Vary-base training.

\textbf{General data}. The processes of training Vary-base follows popular LVLMs, \textit{e.g.}, LLaVA~\cite{llava}, including the pretrain and SFT phases. Different from the LLaVA, we freeze all the vocabulary networks and unfreeze both the input embedding layer and LLM, which is more like the pretrain setting of a pure LLM. We use natural image-text pair data to introduce the general concepts to the Vary-base. The image-text pairs are randomly extracted from LAION-COCO~\cite{schuhmann2021laion} with the amount of 4 million.  In the SFT stage, we use the LLaVA-80k or LLaVA-CC665k~\cite{liu2023improvedllava} along with the train set of DocVQA~\cite{DocVQA} and ChartQA~\cite{masry2022chartqa} as the fine-tuning dataset.

\subsubsection{Conversation format}
When we use the Vicuna-7B as our LLM, the conversation format follows the Vicuna v1~\cite{vicuna}, \textit{i.e.}, USER: <img>"<image>"</img> "texts input" ASSITANT: "texts output" </s>. Due to the low efficiency in the text vocabulary of Vicuna to process Chinese, we choose Qwen-7B~\cite{qwen-chat} as the LLM for Chinese processing. When we use the Qwen-7B, we design the conversation style following the LLaVA-MPT~\cite{team2023introducing,llava}, which can be described as: <|im\_start|>user: <img>"<image>"</img> "texts input"<|im\_end|> <|im\_start|>assistant: "texts output" <|im\_end|>.

% \textbf{Training.} We freeze all the vision encoders' parameters and train others in the Vary-base model for 5 epochs with a batch size of 96. We adopt the AdamW optimizer and the cosine annealing scheduler with an initial learning rate $lr_{init}=9\cdot 10^{-5}$.

\section{Experiments} \label{exp}

\subsection{Datasets and Evaluation Metrics}
We evaluate the proposed Vary on multiple datasets, including 1) a document-level OCR test set we created to explore the performance of dense visual perception; 2) DocVQA~\cite{DocVQA} and ChartQA~\cite{masry2022chartqa} to test the improvement on downstream tasks; 3) MMVet~\cite{yu2023mm} to monitor changes in the general performance of the model.  Our own document test set contains pure OCR and markdown conversion tasks. In a pure OCR task, the test split includes 100 pages in both Chinese and English, which are randomly extracted from arxiv and ebook.  In the markdown conversion task, the test set obtains 200 pages, of which 100 pages contain tables and another 100 pages have mathematical formulas.

We report Normalized Edit Distance~\cite{levenshtein1966binary,blecher2023nougat} and F1-score along with the precision and recall for document parsing. For DocVQA, ChartQA, and MMVet, we use their vanilla metrics for a fair comparison with other LVLMs.

\subsection{Implementation Details}
During the vision vocabulary generating process, we optimize all parameters of Vary-tiny with a batch size of 512 and train the model for 3 epochs. We utilize the AdamW~\cite{AdamW} optimizer and a cosine annealing scheduler~\cite{loshchilov2016sgdr} along with the learning rate of 5e-5 to train Vary-tiny.

In the training stage of the Vary-base, we freeze the weights of both new and vanilla (CLIP-L) vision vocabulary networks and optimize the parameters of input embedding layers and LLM. The initial learning rate is 5e-5 in pretrain while 1e-5 in SFT. Both the pretrain and SFT enjoy a batch size of 256 and an epoch of 1. Other settings are the same as Vary-tiny.

\begin{table*}[h]
\footnotesize
\centering
%\resizebox{1.\linewidth}{!}
\setlength{\tabcolsep}{9pt}
{
\begin{tabular}{llccccc}
\toprule[.9pt]
\multirow{2}{*}{\textbf{Method}}& \multirow{2}{*}{\textbf{Forms}} & \multicolumn{2}{c}{\textbf{Pure Document OCR}} & \multicolumn{3}{c}{\textbf{Markdown Format Conversion}}  \\ 
\cmidrule(rl){3-4} \cmidrule(rl){5-7} &  & Chinese & English & Formula  & Table & Average \\  \midrule
\multirow{3}{*}{Nougat~\cite{blecher2023nougat}}  
& Edit Distance $\downarrow$ &  -- &  0.126 &  0.154 &  0.335 &  0.245  \\ 
& F1-score $\uparrow$ &  -- &  \bf{89.91} &  83.97 & 75.97  &  79.97   \\
& Prediction $\uparrow$ &  -- &  89.12 &  82.47 &  75.21 &  78.84  \\ 
& Recall $\uparrow$ &  -- &  \bf{90.71} &  \bf{85.53} &  \bf{76.74} &  81.14   \\ 
\midrule
\multirow{3}{*}{Vary-tiny} 
& Edit Distance $\downarrow$ & 0.266 &  0.197 &  -- &  -- &  --    \\ 
& F1-score $\uparrow$ & 86.00 &  84.25 &  -- &  -- &  --  \\ 
& Prediction $\uparrow$ & 86.14 &  89.38 &  -- &  -- &  --   \\  
& Recall $\uparrow$ & 85.86 &  79.67 &  -- &  -- &  --  \\ 
\midrule
\multirow{3}{*}{Vary-base} 
& Edit Distance  $\downarrow$ & 0.174 &  \bf{0.106} &  \bf{0.082} &  \bf{0.280} &  0.181    \\ 
& F1-score $\uparrow$         & 87.32 &  88.24      &  \bf{85.94} & \bf{76.26}  &  81.10   \\ 
& Prediction $\uparrow$       & 86.59 &  \bf{90.08} &  \bf{87.06} &  \bf{76.81} &  81.94   \\ 
& Recall $\uparrow$           & 88.06 &  86.47 &  84.84 &  75.71 &  80.28  \\ 
% \midrule
 % \rowcolor{aliceblue!60} Merlin & 7B & 91.24 & 90.98 & 45.6 & 89.23 & 88.87 & 46.77 & 86.23 & 86.18 & 49.63 \\ 
% Merlin & \textbf{91.58} & \textbf{91.66} & 49.38 & \textbf{89.53} & \textbf{89.56} & 50.27 & \textbf{84.10} & \textbf{84.95} & 55.63 \\ 
\bottomrule[.9pt]
\end{tabular}
% \vspace{-.6em}
\caption{Fine-grained text perception compared to Nougat. Vary-tiny is the model based on OPT-125M to generate the vision vocabulary, which enjoys pure OCR ability, including Chinese and English. Vary-base is the model upon Qwen-Chat 7B after scaling up the vision vocabulary, enjoying both pure document OCR and markdown format conversation abilities through prompt control.}
\label{tab:1}
}
% \vspace{-2.em}
\end{table*}

\subsection{Fine-grained Perception Performance}

We measure the fine-grained perception performance of Vary through the dense text recognition ability. As shown in Table~\ref{tab:1}, Vary-tiny gathers both Chinese and English dense OCR ability by the process of vision vocabulary generating. Specifically, it achieves 0.266 and 0.197 edit distance for Chinese and English documents (plain texts) OCR respectively, proving the new vision vocabulary enjoys good fine-grained text encoding capacity.  For Vary-base, it can achieve an on-par performance with nougat~\cite{blecher2023nougat} (a special document parsing model) on  English plain text documents. Besides, with different prompts (\textit{e.g.}, Convert the image to markdown format.), Vary-base can realize the document image-markdown format conversion. It is worth noting that in such a task, Vary-base (with 0.181 edict distance and 81.10\% F1 on math and table average) is better than nougat (with 0.245 edict distance and 79.97\% F1 on average) to some extent, which may be due to the super strong text correction ability of the 7B LLM (Qwen). All the above results indicate that by scaling up the vision vocabulary, the new LVLM can lift its fine-grained perception performance.

% \begin{wrapfigure}{r}{0.5\textwidth}
% \vspace{-5ex}
% 	\centering
% 	\includegraphics[width=0.5\textwidth]{Figs/vary-3.pdf}
% 	\caption{The structure of new vision vocabulary network. We add two convolution layers to convert the output to be similar with CLIP. }
% 	\label{fig3}
% \vspace{-3ex}
% \end{wrapfigure}

% \begin{wraptable}{l}{0.5\textwidth}[!h]
\begin{table}[!h]
        \centering
	\begin{tabular}{lccccc}
        \toprule[.9pt]
        \multirow{3}{*}{\textbf{Method}} & \multicolumn{2}{c}{DocVQA} &\multicolumn{3}{c}{ChartQA} \\
        \cmidrule(rl){2-3}  \cmidrule(rl){4-6}
             & \textbf{val} & \textbf{test} & \textbf{human} & \textbf{augmented} & \textbf{Average}  \\
            \midrule
		Dessurt~\cite{davis2022end}  &  46.5 & 63.2 & -    & -    & - \\
            Donut~\cite{kim2022ocr} &  - & 67.5 & - & - & 41.8\\
            % InstructBLIP~\cite{dai2023instructblip} &  -    & 49.5 & 33.4 & 44.0 & - & 25.6\\
            Pix2Sturct~\cite{lee2023pix2struct} &   - &    72.1 & 30.5 & 81.6 & 56.0 \\
            mPLUG-DocOwl~\cite{ye2023mplug}  & -  & 62.2 & - & - & 57.4 \\
            Matcha~\cite{liu2022matcha} &  - & - & 38.2 & \underline{90.2} & 64.2\\
            Qwen-VL~\cite{qwen}   &  - & 65.1 & - & - & 65.7 \\
            \midrule
            Vary-base (80k)  &  \underline{78.2} & 76.3 & 43.2 & 87.3 & 65.3\\
            Vary-base (665k)  &  78.1 & 76.3 & \underline{43.8} & 88.3 & \underline{66.1} \\
        % \midrule    
        % \rowcolor{aliceblue!60} \textbf{Merlin} (Ours)  &  \underline{60.5} & \bf50.4 &	\bf66.2 & \bf65.5 & \bf34.9 \\
        \bottomrule[.9pt]
	\end{tabular}
        \setlength{\abovecaptionskip}{0.2cm}
       \caption{Comparison with popular methods on DocVQA and ChartQA. 80k represents that the SFT data is LLaVA-80k while 665k is the LLaVA-CC665k. The metric of DocVQA is ANLS while the ChartQA is relaxed accuracy following their vanilla papers.}
        \label{tab:2}
        % \vspace{-5mm}
\end{table}
% \end{wraptable}

\begin{table}[!h]
        \centering
	\begin{tabular}{lccccccc}
        \toprule[.9pt]
        \multirow{3}{*}{\textbf{Method}} & \multicolumn{7}{c}{MM-Vet}\\
        \cmidrule(rl){2-8}
             & \textbf{Rec} & \textbf{OCR} & \textbf{Know} & \textbf{Gen} & \textbf{Spat} & \textbf{Math} & \textbf{Total} \\
            \midrule
		BLIP-2~\cite{BLIP2}   &  27.5 & 11.1 & 11.8 & 7.0 & 16.2 & 5.8 & 22.4 \\
            LLaVA-7B~\cite{llava} & 28.0 & 17.1 & 16.3 & 18.9 & 21.2 & \underline{11.5} & 23.8\\
            % InstructBLIP~\cite{dai2023instructblip} &  -    & 49.5 & 33.4 & 44.0 & - & 25.6\\
            MiniGPT-4~\cite{minigpt4}  & 29.9 & 16.1 & 20.4 & 22.1 & 22.2 & 3.8 & 24.4 \\
            Otter~\cite{li2023otter} & 27.3 & 17.8 & 14.2 & 13.8 & 24.4 & 3.8 & 24.7 \\
            OpenFlamingo~\cite{Flamingo} & 28.7 & 16.7 & 16.4 & 13.1 & 21.0 & 7.7 & 24.8\\
            LLaVA-13B~\cite{llava} & \underline{39.2} & 22.7 &\underline{26.5} &\underline{29.3} & 29.6 & 7.7 & 32.9 \\
            LLaVA1.5-7B~\cite{liu2023improvedllava} & - & - & - & -& - & - & 30.5 \\
            \midrule
            Vary-base (vicuna7B) (665k)  &  38.7 & 22.0 & 23.6 & 24.1 & 29.6 & 7.7    & 32.9 \\
            Vary-base (qwen7B) (80k)  &  38.9 & \underline{30.1} & 22.4 & 21.7 & \underline{34.3} & 7.7    & \underline{36.2} \\
        % \midrule    
        % \rowcolor{aliceblue!60} \textbf{Merlin} (Ours)  &  \underline{60.5} & \bf50.4 &	\bf66.2 & \bf65.5 & \bf34.9 \\
        \bottomrule[.9pt]
	\end{tabular}
        \setlength{\abovecaptionskip}{0.2cm}
       \caption{Comparison with popular methods on MMVet. The abbreviations represent: Rec: Recognition; Know: Knowledge; Gen: Language generation; Spat: Spatial awareness.}
        \label{tab:3}
        \vspace{-5mm}
\end{table}

\subsection{Downstream Task Performance}

We test the performance improvement on downstream VQA tasks with DocVQA~\cite{DocVQA} and ChartQA~\cite{masry2022chartqa}. We use the addition prompt: "Answer the following questions using a single word or phrase:"~\cite{liu2023improvedllava} to allow the model to output short and precise answers. As shown in Table~\ref{tab:2}, Vary-base (with Qwen-7B as LLM) can achieve 78.2\% (test) and 76.3\% (val) ANLS on DocVQA upon LLaVA-80k~\cite{llava} SFT data. With LLaVA-665k~\cite{liu2023improvedllava} data for SFT, Vary-base can reach 66.1\% average performance on ChartQA. The performance on both two challenging downstream tasks is comparable to or even better than Qwen-VL~\cite{Qwen-VL}, demonstrating the proposed vision vocabulary scaling-up method is also promising for downstream.

\subsection{General Performance}
We monitor the general performance of Vary through MMVet~\cite{yu2023mm} benchmark.  As shown in table~\ref{tab:3}, with the same LLM (Vicuna-7B) and SFT data (LLaVA-CC665k), Vary lifts 2.4\% (32.9\% vs. 30.5\%) of the total metric than LLaVA-1.5, proving that our data and training strategy do not hurt the model's general ability. Besides, Vary with Qwen-7B and LLaVA-80k can achieve 36.2\% performance, further demonstrating the effectiveness of our vision vocabulary scaling-up manner.

% We report the following metrics on our test set.

% \textbf{Edit distance} The edit distance\cite{Edit-distance}, measures the number of character manipulations (insertions, deletions, substitutions) it takes to get from one string to another. In this work, we consider the normalized edit distance, where we divide by the total number of characters. A smaller edit distance is better.

% \textbf{F-measure} F-measure reports the characters recognized precision and recall and calculates the F1-score. The indicators in F-measure bigger are better.

\section{Conclusion}
\label{discussion}
This paper highlights that scaling up the vocabulary in the visual branch for an LVLM is quite significant and we successfully devise a simple method to prove such a claim. According to the experiments, the provided model, Vary, achieves promising scores in multiple tasks, which is mainly profited by the new vocabulary we generated. Despite the satisfactory performance of Vary, we believe that how to effectively scale up the visual vocabulary still enjoys much improvement rooms, especially compared to the mature and relatively simple means of expanding text vocabulary. We hope that the useful and efficient design of Vary will attract more research attention to such a direction.

\section{Appendix}
In this appendix, we present the output results of our model to provide a more intuitive understanding of its performance.

% \subsection{Visualization Results}

% We present the output results of our model to provide a more intuitive understanding of its performance.
\begin{figure}[h]
\centering
\includegraphics[width=1.0\textwidth]{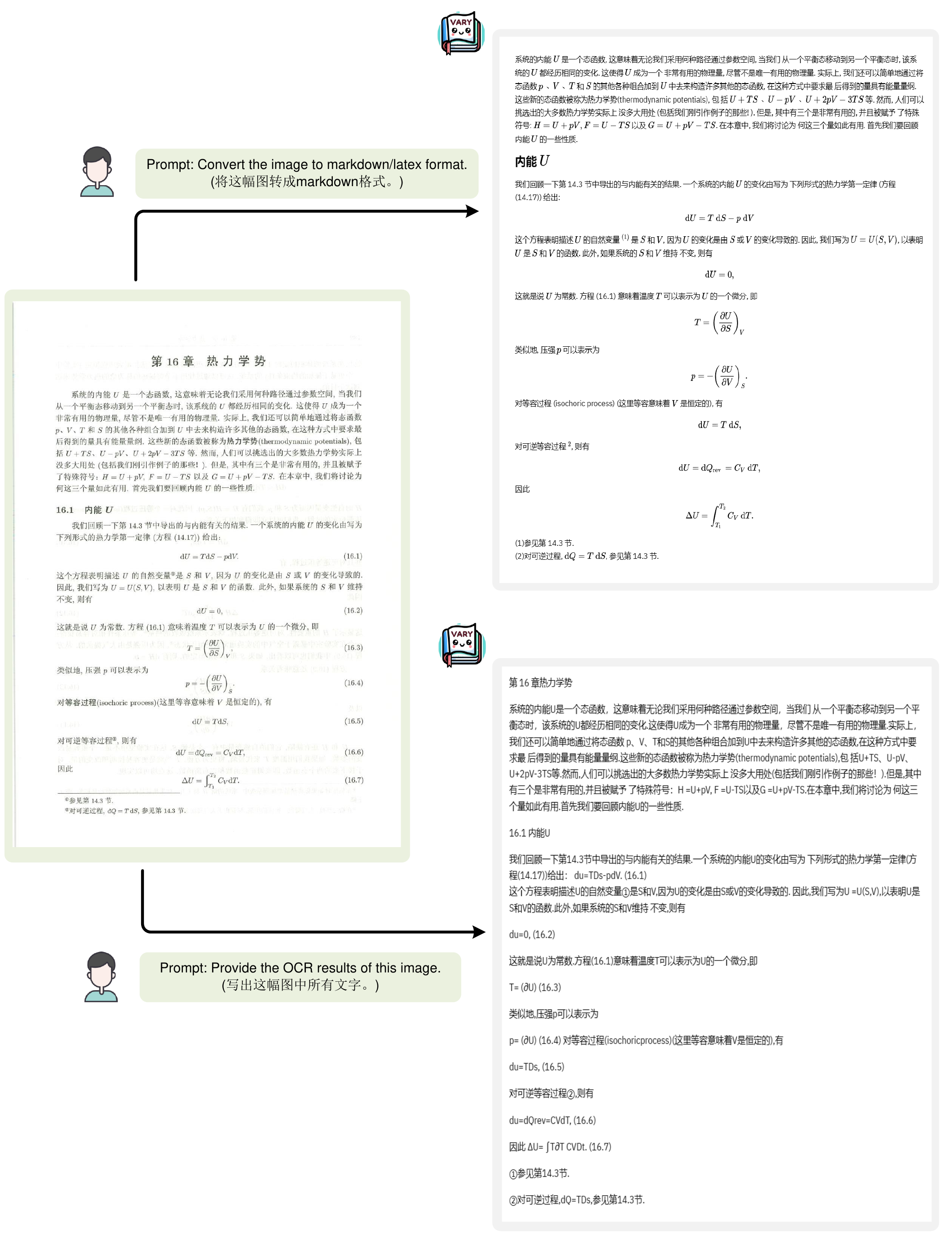}
\caption{Instruction following ability of Vary-base to excel markdown conversion or pure OCR. Vary-base can control the output format for a document image input upon the user's prompts.}
\label{figa1}
\end{figure}

\begin{figure}[h]
\centering
\includegraphics[width=1.0\textwidth]{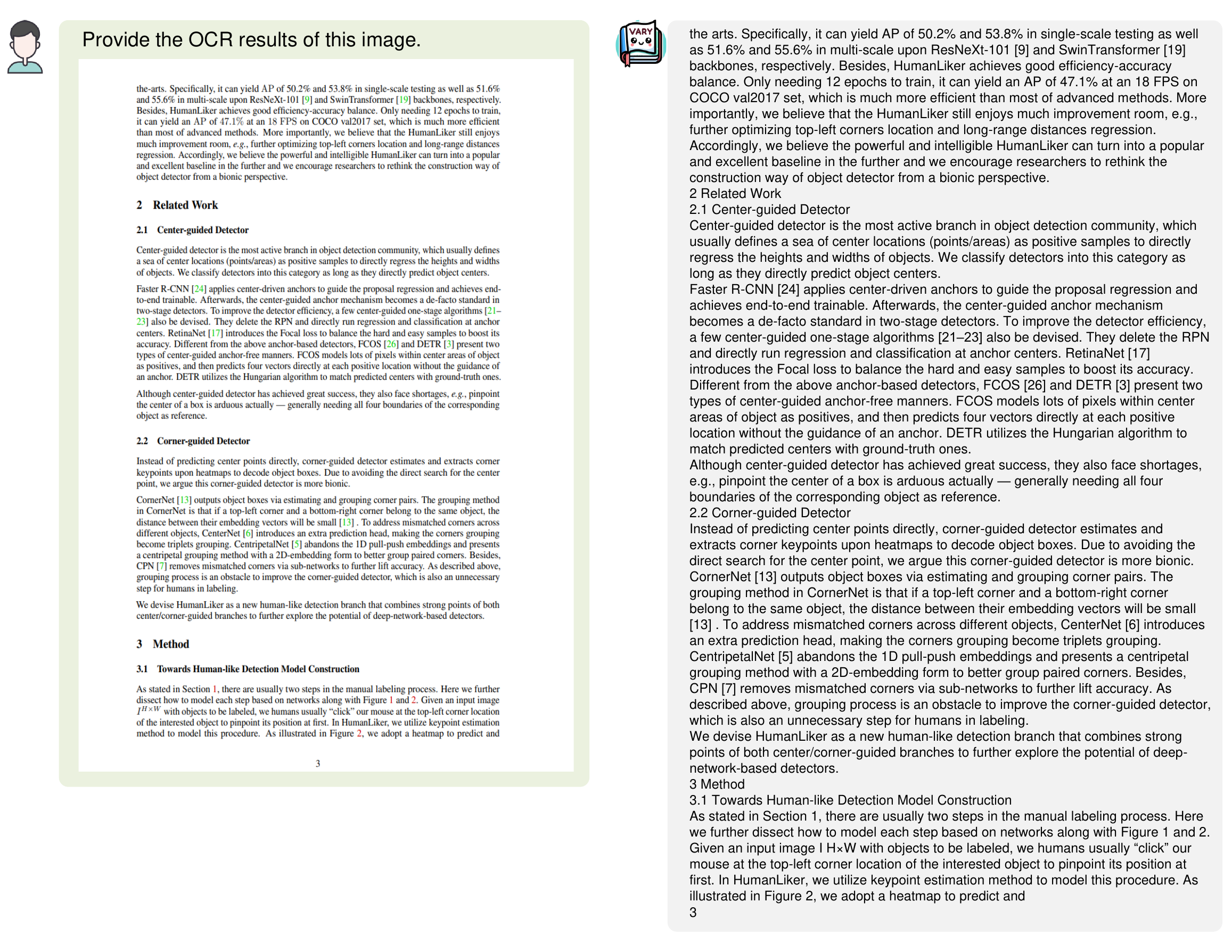}
\caption{Fine-grained visual perception ability of Vary-base on English document dense OCR. This image is the page 3 of ~\cite{wei2022humanliker}.}
\label{figa2}
\end{figure}

\vspace{10mm}

\begin{figure}[!h]
\centering
\includegraphics[width=1.0\textwidth]{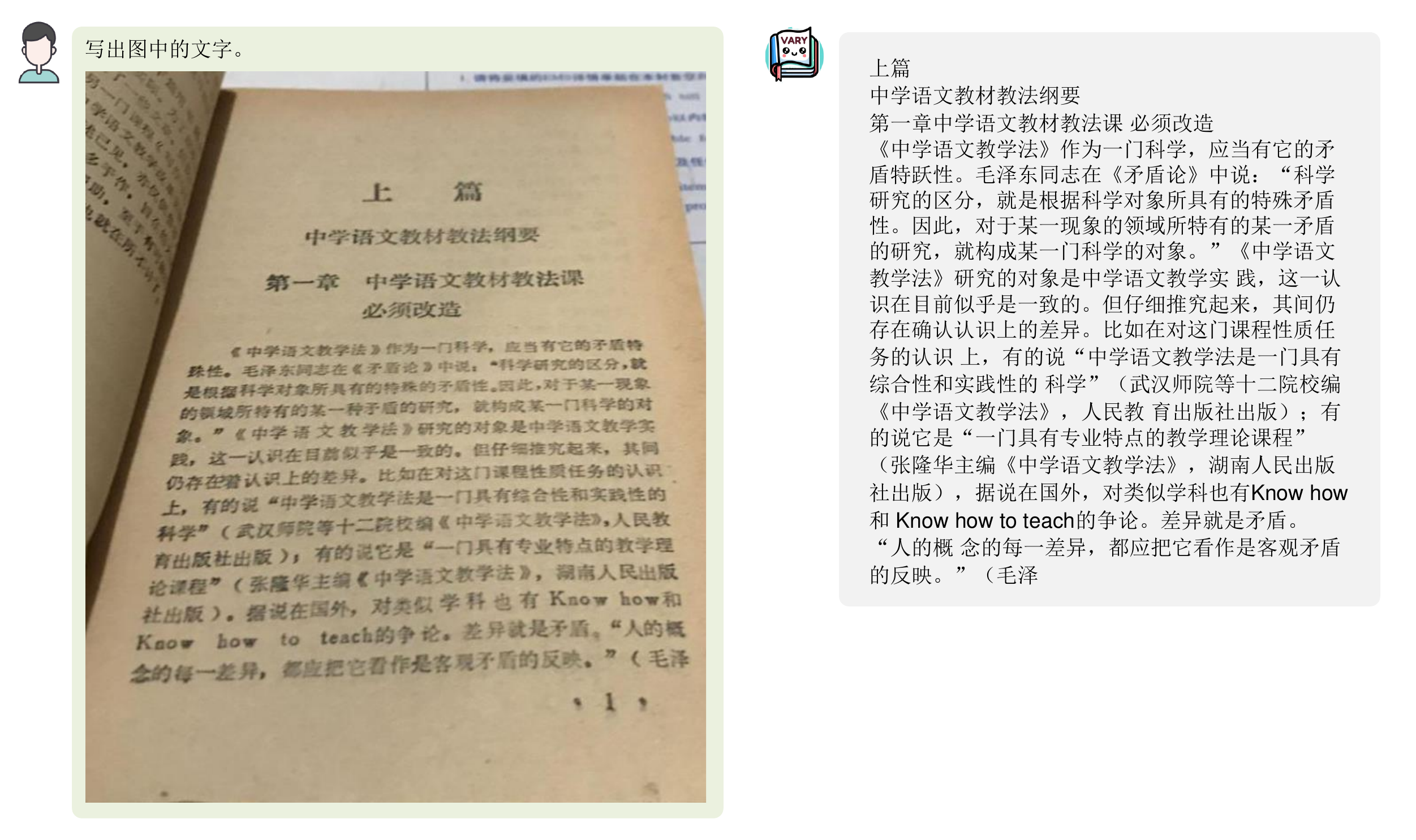}
\caption{Fine-grained visual perception ability of Vary-base on Chinese book dense OCR. This image is from the Internet.}
\label{figa3}
\end{figure}

\vspace{30mm}

\begin{figure}[!h]
\centering
\includegraphics[width=1.0\textwidth]{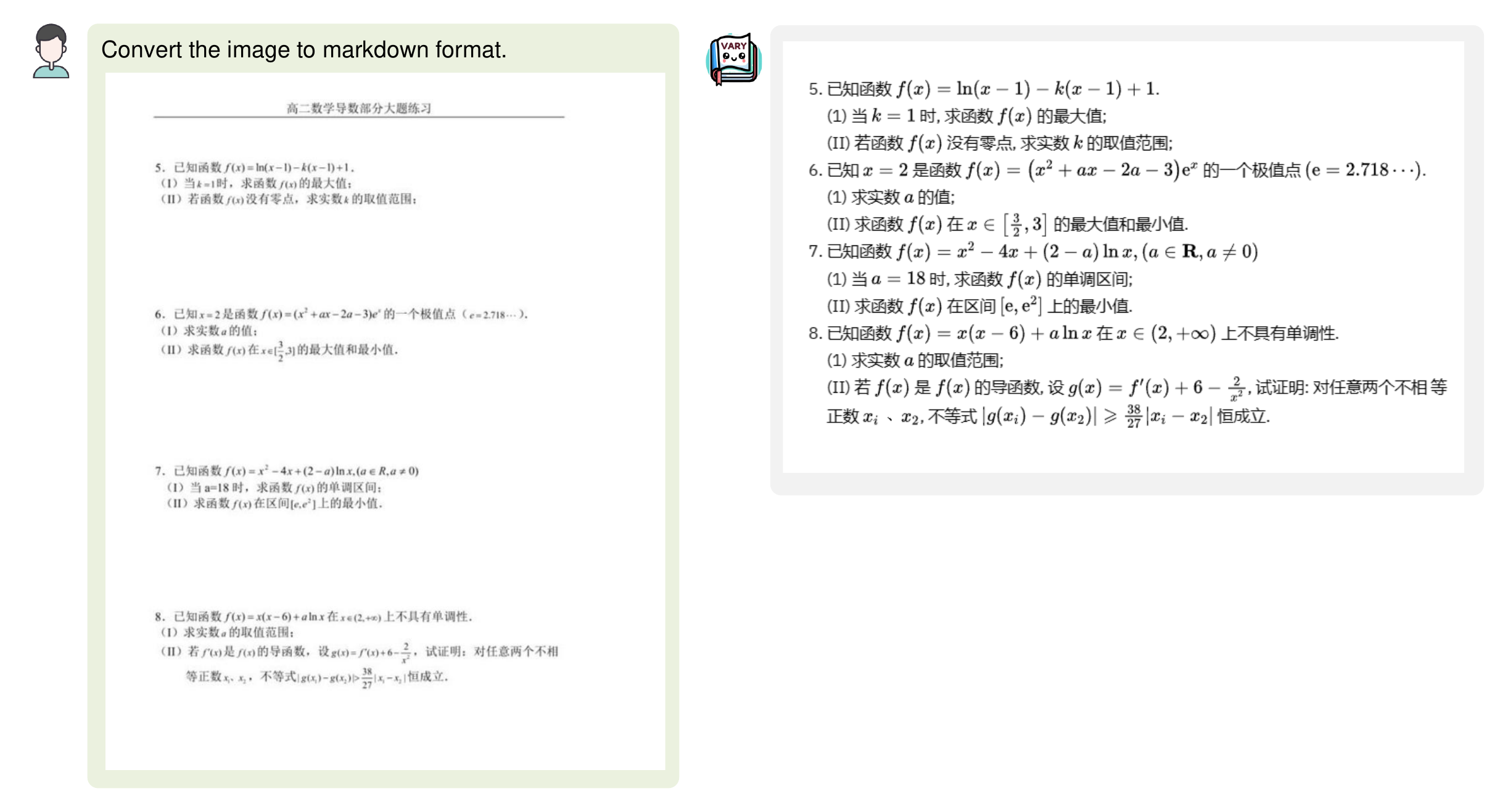}
\caption{Markdown/Latex format conversion ability (on math formula) of Vary-base. This image is from the Internet.}
\label{figa4}
\end{figure}

\vspace{20mm}

\begin{figure}[!h]
\centering
\includegraphics[width=1.0\textwidth]{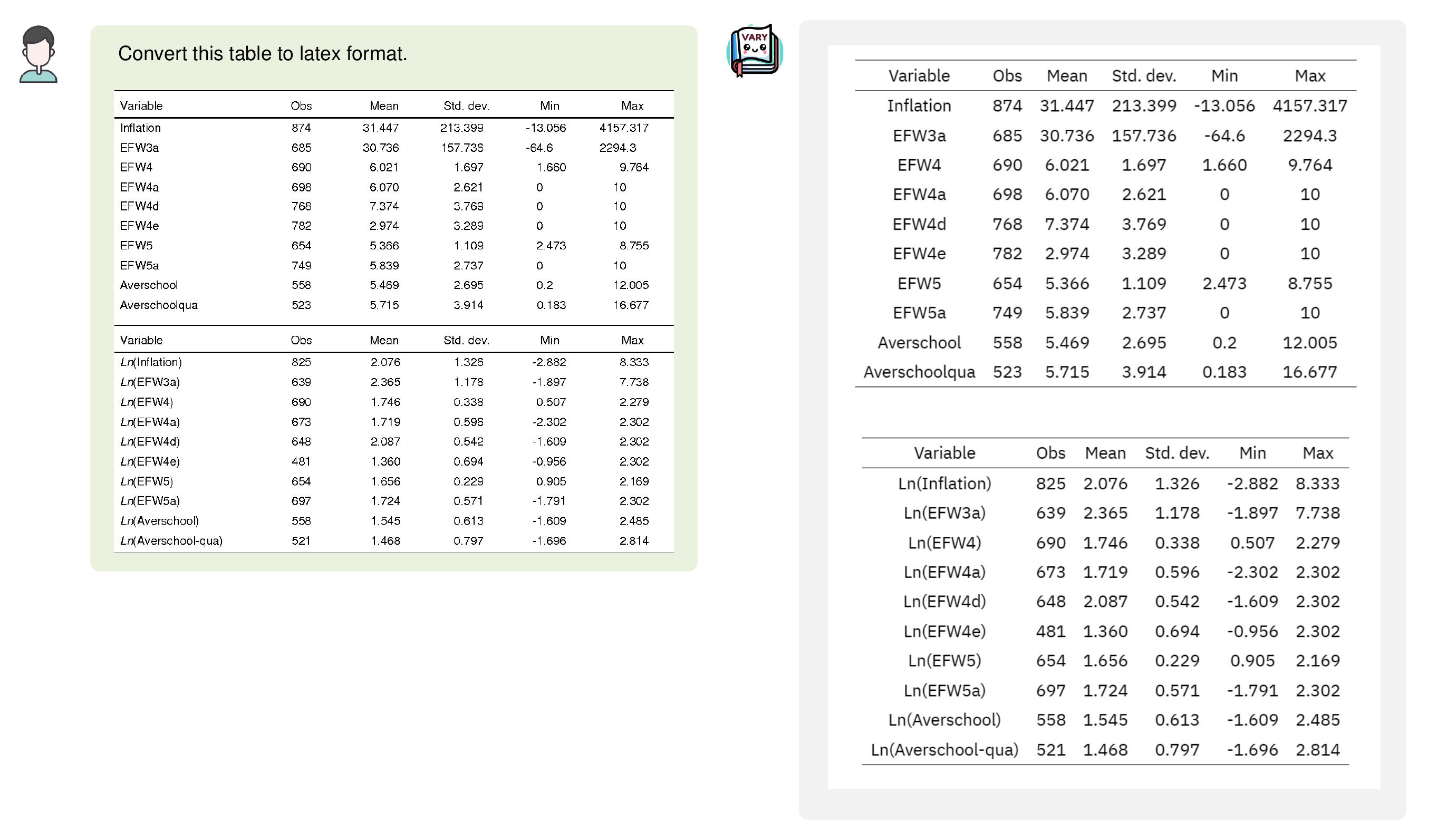}
\caption{Markdown/Latex format conversion ability (on the table) of Vary-base.The images are from the Internet.}
\label{figa5}
\end{figure}

% \vspace{45mm}

\begin{figure}[!t]
\centering
\includegraphics[width=0.85\textwidth]{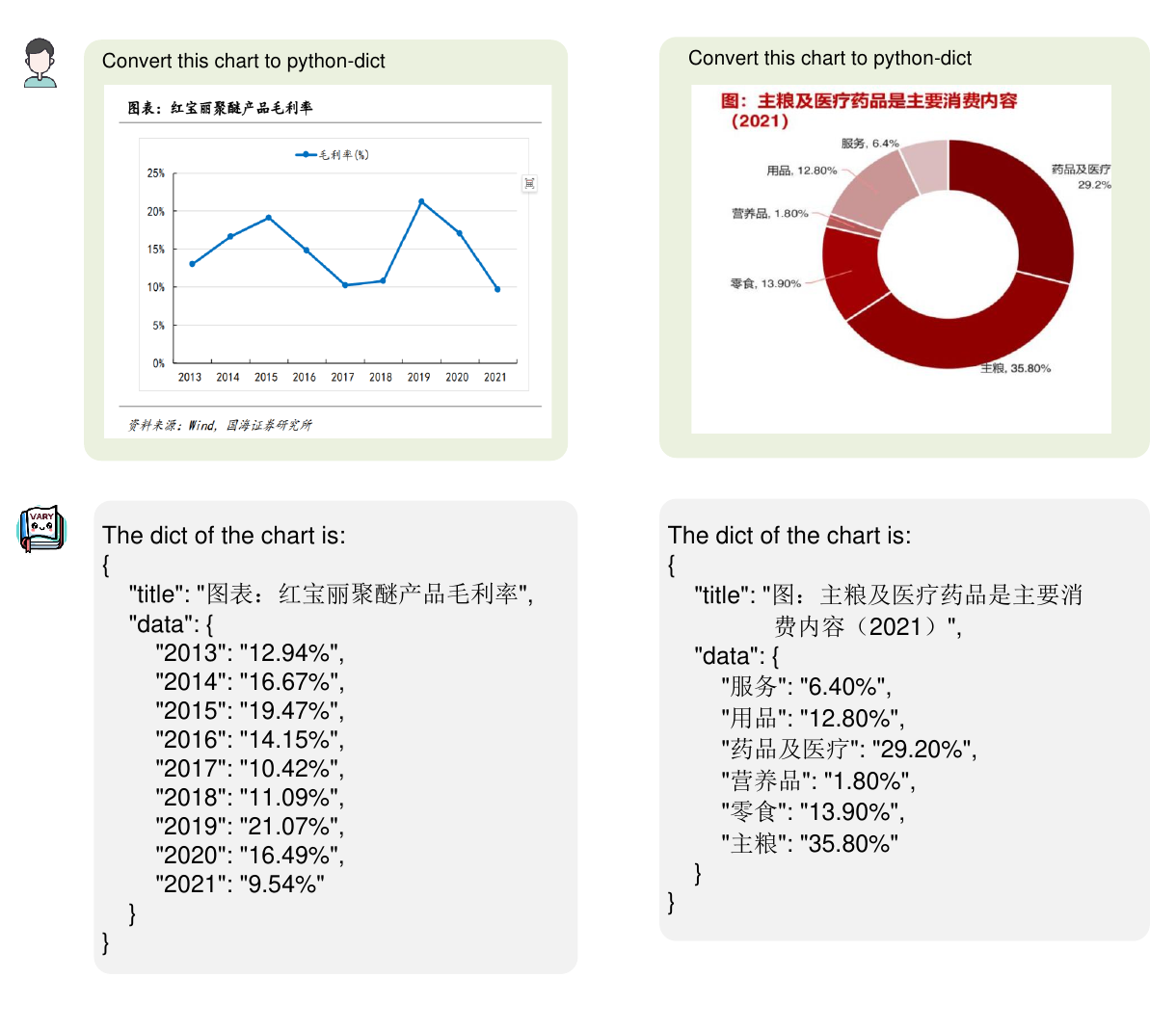}
\caption{Chart understanding (Chinese) of Vary-base. The images are from the Internet.}
\label{figa5}
\end{figure}

\vspace{15mm}

\begin{figure}[!h]
\centering
\includegraphics[width=0.92\textwidth]{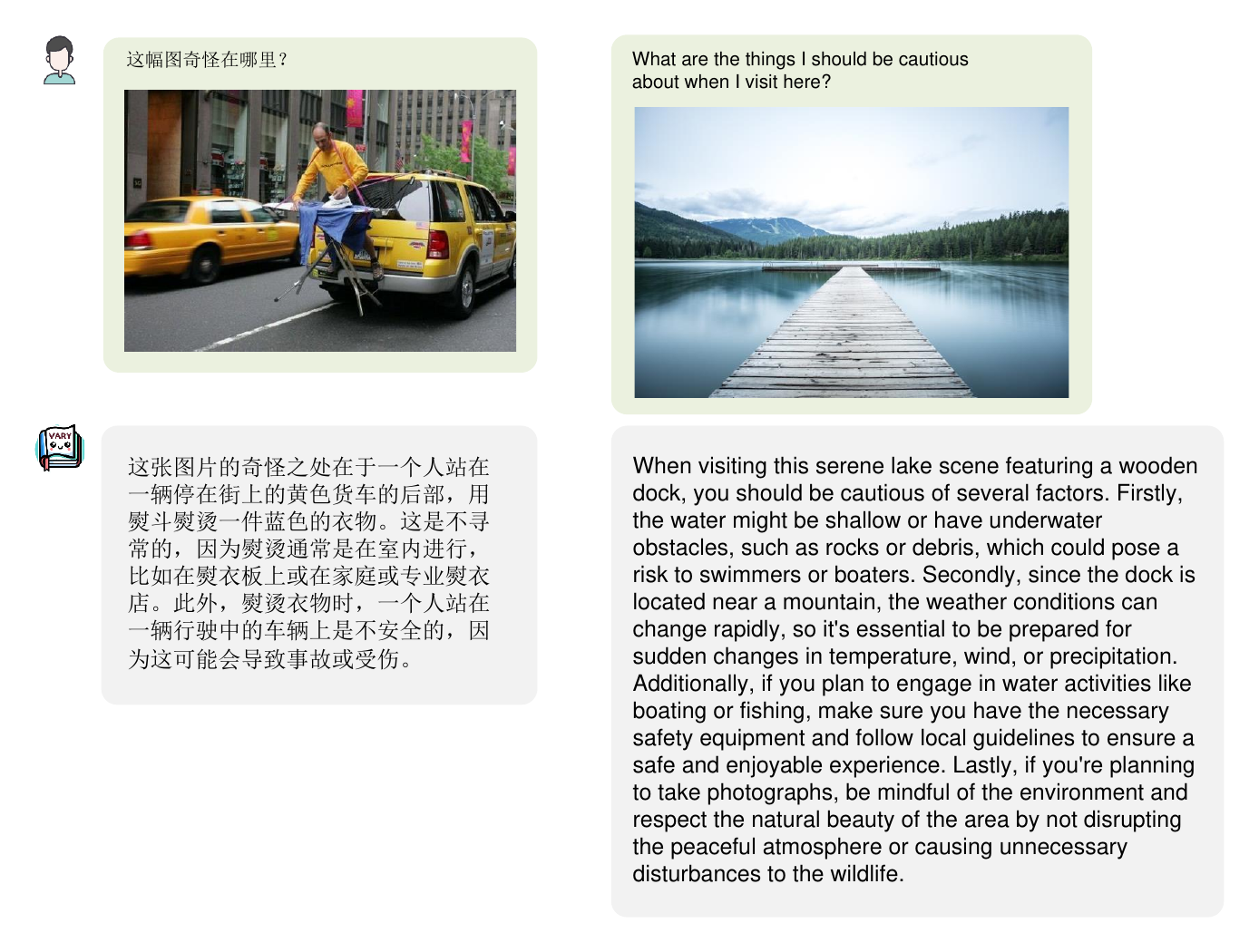}
\caption{General performance of Vary-base. The images are from LLaVA~\cite{llava} samples.}
\label{figa5}
\end{figure}

% \subsection{Rendered Data}
% In this section, we show more rendered data used in Vary.

\newpage

%%% END INSTRUCTIONS %%%
{
\small
\bibliographystyle{splncs04}
\bibliography{egbib}
}

%%%%%%%%%%%%%%%%%%%%%%%%%%%%%%%%%%%%%%%%%%%%%%%%%%%%%%%%%%%%
% Optionally include extra information (complete proofs, additional experiments and plots) in the appendix.
% This section will often be part of the supplemental material.

\end{document}